\documentclass{llncs}  

\usepackage{graphicx} 
\usepackage[T1]{fontenc}
\usepackage{amsmath}
\usepackage[english]{babel}
\usepackage{array,tabularx}
\usepackage{booktabs} 
\usepackage{makeidx}  
\usepackage{xparse}
\usepackage{moredefs}
\usepackage{lips}
\usepackage{epstopdf} 
\usepackage{color}
\usepackage{csquotes}
\usepackage{xcolor}
\usepackage{times}
\usepackage[hidelinks]{hyperref}
\usepackage[nolist,nohyperlinks]{acronym}
\usepackage{comment}
\usepackage{paralist}
\usepackage{cleveref}
\usepackage{wrapfig}
\usepackage{multirow}
\usepackage{microtype}
\usepackage{listings}
\usepackage[super]{nth}
\usepackage{natbib}
\usepackage{siunitx}
\usepackage{tabularx}

\usepackage{float}
\floatstyle{plaintop}
\restylefloat{table}

\usepackage[style=base,labelfont=bf,labelsep=period,tableposition=top,font=small]{caption}
\makeatletter
\renewcommand\@biblabel[1]{#1.}
\makeatother
\addto\captionsenglish{}

\usepackage{fancyhdr}
\fancypagestyle{WI_footer}{\fancyhf{}\fancyfoot[L]{\color{black}\scriptsize 20th International Conference on Wirtschaftsinformatik\\September 2025, Münster, Germany}}

\crefformat{footnote}{#2\footnotemark[#1]#3}
\expandafter\def\expandafter\UrlBreaks\expandafter{\UrlBreaks
  \do\a\do\b\do\c\do\d\do\e\do\f\do\g\do\h\do\i\do\j%
  \do\k\do\l\do\m\do\n\do\o\do\p\do\q\do\r\do\s\do\t%
  \do\u\do\v\do\w\do\x\do\y\do\z\do\A\do\B\do\C\do\D%
  \do\E\do\F\do\G\do\H\do\I\do\J\do\K\do\L\do\M\do\N%
  \do\O\do\P\do\Q\do\R\do\S\do\T\do\U\do\V\do\W\do\X%
  \do\Y\do\Z}

\newcolumntype{L}[1]{>{\raggedright\arraybackslash}p{#1}}   
\newcolumntype{C}[1]{>{\centering\arraybackslash}p{#1}}     
\newcolumntype{R}[1]{>{\raggedleft\arraybackslash}p{#1}}    

\begin{document}
\frontmatter          

\mainmatter              

\title{Analyzing German Parliamentary Speeches: A Machine Learning Approach for Topic and Sentiment Classification}

\subtitle{Research Paper} 
\author{Lukas Pätz\inst{1} \and
Moritz Beyer\inst{1} \and
Jannik Späth\inst{1} \and
Lasse Bohlen\inst{1,2} \and 
Patrick Zschech\inst{1,2} \and 
Mathias Kraus\inst{3} \and
Julian Rosenberger\inst{3}}

\institute {Leipzig University, Leipzig, Germany \\ \email{\{lukas.paetz, zu66evit, j.spaeth\}@studserv.uni-leipzig.de} 
\and TU Dresden, Dresden, Germany \\ \email{\{lasse.bohlen, patrick.zschech\}@tu-dresden.de} 
\and University of Regensburg, Regensburg, Germany \\ \email{\{mathias.kraus, julian.rosenberger\}@ur.de} \\ 
}

\maketitle
\setcounter{footnote}{0}

\begin{abstract}
This study investigates political discourse in the German parliament, the Bundestag, by analyzing approximately 28,000 parliamentary speeches from the last five years. Two machine learning models for topic and sentiment classification were developed and trained on a manually labeled dataset. The models showed strong classification performance, achieving an area under the receiver operating characteristic curve (AUROC) of 0.94 for topic classification (average across topics) and 0.89 for sentiment classification. Both models were applied to assess topic trends and sentiment distributions across political parties and over time. The analysis reveals remarkable relationships between parties and their role in parliament. In particular, a change in style can be observed for parties moving from government to opposition. While ideological positions matter, governing responsibilities also shape discourse. The analysis directly addresses key questions about the evolution of topics, sentiment dynamics, and party-specific discourse strategies in the Bundestag.\\

{\bfseries Keywords:} Natural Language Processing, German Parliamentary, Discourse Analysis, Bundestag

\end{abstract}

\thispagestyle{WI_footer}


\section{Introduction}
\label{sec:introduction}

In today’s digital age, political discussions primarily take place on social media and in televised debates, where public discourse is often shaped by rapid exchanges of opinions \citep{emmer2017soziale}. However, the parliament remains the primary institution where formal political decision-making occurs and legislative processes unfold. While digital platforms amplify public opinion, it is in the parliament where policies are debated, negotiated and enacted -- making it a crucial forum for understanding political discourse. Germany, with its population of over 80 million, is not only one of the largest economies in the world \citep{sahoo2024drivers}, but also a country where political decisions have far-reaching national and international consequences. The German parliament, known as the Bundestag, plays an important role in shaping the nation's political discourse and serves as a center for critical debates on various societal issues, and it facilitates rigorous discussions on these topics \citep{deutscher_bundestag_arbeit_2025}.

Elected representatives utilize this platform to articulate public concerns, commend policy achievements, and critically evaluate governmental actions. During the legislative period between 2021 and 2025, the 20th German Bundestag was the largest freely elected national parliament in the world with 733 members. Each year, approximately 80 plenary sessions are held, during which several thousand speeches on diverse topics are delivered \citep{deutscher_bundestag_arbeit_2025}. 
These public debates on key political issues provide a rich foundation of textual data for analysis.

In recent years, a perceived divide between the political elite and the general population in Germany has grown. The German phrase "die da oben" (those up there), often used metaphorically to refer to political decision-makers such as members of parliament, has regained prominence as a symbol of growing public distrust \citep{Hamann2016}. 
Studies and surveys highlight a growing skepticism among the population towards political institutions and their actors. According to the  \citet{korber_stiftung_demokratie_2024}, 54\% of German respondents reported having little or no trust in democracy. In 2019, this figure stood at just 30\%, highlighting a dramatic and accelerating loss of confidence in Germany’s democratic system over the past years \citep{Koerber_stiftung_demokratie_2023b}. This development is confirmed by the \citet{EuropeanCommission2024}, which shows that by early 2024, only 33\% of EU citizens expressed trust in their national governments. Germany, which had traditionally ranked above the EU average, is now among the countries with the sharpest declines in trust. These developments reflect a broader demand for increased transparency, accountability, and comprehensibility in political processes. 

Many studies have examined the influence of the media and the public's changing tone on this development \citep{boulianne2020twenty, lorenz2023systematic}. While media coverage and social media trends often highlight selective aspects of political discourse, Bundestag debates offer unfiltered insights into policy discussions. Analyzing parliamentary discourse on a large scale reveals the priorities, ideologies, and rhetorical strategies of political actors. From an academic perspective, this enhances the understanding of institutional communication, and from a societal perspective, it supports transparency and accountability by showing how parties address key issues over time.

Modern Natural Language Processing (NLP) techniques offer the possibility of analyzing these debates on a large scale, identifying dominant topics and assessing their tone \citep{abercrombie2020sentiment}.

Understanding which topics dominate Bundestag debates, how they are framed by different political parties, and whether discussions are characterized by support or criticism is essential for assessing the dynamics of political communication. Our study investigates how topics and sentiments in Bundestag speeches evolved between 2019 and 2024. The overarching research question guiding this work is:

\medskip
\noindent\textbf{\textit{Research Question:}} \emph{What patterns in the evolution of topics and sentiments can be identified in Bundestag speeches using NLP techniques?}
\medskip

\newpage 
\noindent To explore this question, our analysis focuses on three key aspects:
\begin{enumerate}
    \item Which topics were most prominent, and how did their relevance shift over time?
    \item How did the sentiment of speeches change, particularly with shifts between government and opposition roles?
    \item Did parties use different sentiments when addressing the topics they emphasized?
\end{enumerate}

To address our research questions, we apply machine learning-based classification techniques to Bundestag speeches from two complementary perspectives. First, topic classification allows us to analyze which issues dominate parliamentary debates, how their prominence shifts over time, and how parties differ in their thematic focus and ideological framing. Second, sentiment classification enables us to explore how parties express positive or negative attitudes toward these topics, revealing patterns in political tone, framing, and changes in sentiment across time and party lines.

It is important to emphasize that this study does not aim to endorse any political position. Instead, the goal is to provide tools for objectively analyzing the political discourse in Germany using quantitative data. By doing so, this study seeks to contribute to a deeper understanding of parliamentary discussions.
To support further research and ensure transparency, the full codebase and all associated data are also made available.\footnote{\href{https://github.com/ptzlukas/german-parliamentary-speech-classifier}{https://github.com/ptzlukas/german-parliamentary-speech-classifier}}

The remainder of this paper is structured as follows. Section \ref{sec:related} provides a review of relevant literature, thereby establishing the theoretical foundation for the methodology employed and situating the work within existing research. Section~\ref{sec:methods} outlines the steps involved in generating, annotating and developing the model, as well as external validation. This is followed by a brief description of how the models were applied to the full speech dataset for trend analysis. Section~\ref{sec:results} presents the empirical findings, which are subsequently discussed in Section~\ref{sec:discussion}, where we examine their broader implications, acknowledge the limitations of the study, and propose directions for future research.

\section{Related Work}
\label{sec:related}

Transcripts of legislative debates provide rich material for analyzing the positions and rhetorical strategies of political actors and their parties on key societal issues. In the context of the German Bundestag, such transcripts offer unique insights into how political discourse is structured within a formal legislative setting. While interest in parliamentary speech has grown across disciplines, the Bundestag serves as a particularly valuable case due to its institutional transparency, large volume of data, and diverse party landscape \citep{laver2003extracting}. Political and social scientists, who have traditionally relied on expert coding for analyzing such texts, are increasingly adopting computational approaches, enabling a more systematic analysis of political positions \citep{abercrombie2020sentiment}. These advancements allow for a more data-driven examination of parliamentary discourse and ideological patterns \citep{rauh2018validating}.

Building on these computational advances, text classification, which is a foundational task in both manual and automated text analysis \citep{debortoli2016text, hofmann2024hate}, emerges as a key method for analyzing political discourse. In the context of political discourse, topic and sentiment classification are particularly relevant. 

\emph{Topic classification} seeks to assign textual content to specific thematic categories or topics \citep{nasution2024chatgpt}. Topics represent the primary subject matter or prevailing themes \citep{koltsova2013mapping} within a given text. Often those texts are derived from sources such as social media platforms, news articles, or consumer behavior data \citep{akter2016does}. These topics can either be predefined or derived from a pool of texts. This process of topic discovery can be carried out using a range of computational approaches, including clustering-based methods \citep{ogunleye2023comparison} and machine learning techniques \citep{shah2021mining}, which facilitate the identification and monitoring of emerging trends \citep{kim2016topic}.

\emph{Sentiment classification} is a specific form of text classification that categorizes texts based on the emotional or attitudinal orientation of expressed opinions \citep{devika2016sentiment}. It can be performed through various approaches, including the use of lexicon-based methods that count emotionally charged words, the application of conventional machine learning models such as logistic regression trained on word frequency features or the deployment of advanced neural networks \citep{stine2019sentiment}. The rapid expansion of social media has significantly amplified the accessibility of public opinions and sentiments, thereby establishing sentiment classification as an essential tool for understanding public attitudes across diverse fields including business, politics, and beyond \citep{tan2023survey}. 

While topic classification aims to uncover latent thematic structures within document collections \citep{EickhoffNeuss2017}, sentiment classification is used to extract opinions and attitudes from texts, including those from social media, blogs, or parliamentary debates \citep{storey2022ontology}. Recent advances focus on finer-grained approaches that account for the multi-aspect nature of sentiment -- where a single text can express different sentiments towards various entities or subjects -- which is especially pertinent for analyzing complex political speech \citep{lu2011multi}. 

Over the past decade, interest in parliamentary text corpora has grown steadily, resulting in a series of national projects. One of the first was "DutchParl", introduced by \citet{marx2010dutchparl}, covering debates from the Netherlands, Flanders, and Belgium. Shortly thereafter, \citet{jakubicek2010czechparl} released "CzechParl", based on plenary session transcripts from the Czech parliament. These initiatives were followed by corpora for Finland \citep{mansikkaniemi2017automatic}, Greece \citep{dritsa2022greek}, Poland \citep{ogrodniczuk2018polish}, Norway \citep{fiva2025norwegian}, and Turkey \citep{gungor2018corpus}, illustrating the increasing international interest in the systematic analysis of legislative discourse.

In recent years, various projects have aimed to develop a comprehensive corpus encompassing the parliamentary debates of the German Bundestag. \citet{barbaresi2018corpus} compiles speeches delivered by the German presidents, the Bundestag presidents, the German chancellors, and the foreign ministers from the years 1982 to 2018. Additionally, several initiatives have focused on integrating parliamentary records from multiple countries. "ParlSpeech V2" provides parliamentary protocols from the national parliaments of Austria, Germany, Denmark, and other countries over various legislative periods \citep{rauh2020parlspeech}, while "GerParCor" focuses on protocols and speeches from the national parliaments of Austria, Switzerland, Liechtenstein, and Germany \citep{abrami2022german}. Moreover, \citet{richter2023open} present a machine-readable dataset of speeches from German Bundestag sessions, covering nearly 900,000 speeches from 1949 to 2022. The project provides a structured, maintainable data format enriched with political context, enabling advanced text analysis and supporting diverse political science research.

Building on this existing body of work, the following section details the specific methodological approach we developed to systematically analyze discourse patterns in the Bundestag and address our research questions.

\section{Methods}
\label{sec:methods}

\subsection{Analytical Approach}
\label{sec:analytical_approach}

Our study uses a two-step supervised machine learning approach to analyze a large corpus of parliamentary speeches from the German Bundestag. First, topic and sentiment classifiers were trained and evaluated on a manually labeled subset, with domain-specific annotations ensuring high-quality training data. This allowed the models to capture nuances of political communication and party behavior. In the second step, the trained models were applied to the full dataset of approximately 28,000 speeches by 963 speakers across seven parties. The resulting thematic and sentiment predictions were integrated with metadata (e.g., speaker, party, date), creating a comprehensive annotated dataset.

Unlike studies using generic labels, this approach ensures contextual relevance through annotations tailored to the parliamentary domain. Combining manual labeling with scalable machine learning enables systematic, large-scale analysis of discourse patterns. The resulting dataset offers an empirical basis for examining trends in political language and behavior across parties and over time.

\subsection{Dataset Generation and Annotation}
\label{sec:dataset_annotation}

Supervised machine learning in NLP involves training algorithms on labeled text data so they can learn patterns such as topics or sentiment and accurately classify or predict similar patterns in new, unseen texts \citep{singh2016review}. To create such a labeled dataset we used the Documentation and Information System for Parliamentary Materials (DIP)\footnote{\href{https://search.dip.bundestag.de/api/v1/swagger-ui/}{https://search.dip.bundestag.de/api/v1/swagger-ui/}} interface to download Bundestag plenary meeting minutes from October 2019 to October 2024, resulting in 393 downloaded plenary protocols. Subsequently, individual speeches and relevant fields were extracted from those protocols using regular expressions to distinguish speeches by speaker name, party affiliation, and position because multiple speeches are collected within one plenary protocol \citep{chapman_exploring_2016}.

Data preprocessing involved noise removal, tokenization, lowercase conversion, lemmatization, and partly speech-based stopword removal (via SpaCy) \citep{smelyakov_effectiveness_2020}. A custom stopword list for German parliamentary debates was used\footnote{\href{https://github.com/ptzlukas/german-parliamentary-speech-classifier/blob/main/resources/stopwords.txt}{https://github.com/ptzlukas/german-parliamentary-speech-classifier/blob/main/resources/stopwords.txt}} for topic classification but omitted for sentiment classification to preserve model performance.

Manual labeling was used to create a ground truth, which is essential for effective sentiment and topic classification. The authors conducted the labeling manually for both topic and sentiment based on a randomly selected subset of speeches. Due to the thematic complexity of many of the speeches, the selected samples were cross-checked by other group members to ensure consistency and reduce subjectivity. The number and scope of topic labels were guided by insights from linear discriminant analysis based topic classification \citep{blei2003latent}, which revealed coherent thematic clusters. These clusters were then interpreted and consolidated in a way that reflects the committee structure of the German Bundestag,\footnote{\href{https://www.bundestag.de/en/committees}{https://www.bundestag.de/en/committees}} ensuring both statistical grounding and political relevance. The procedure results in six categories: Environment, Social Affairs and Education, Economy and Finance, Foreign and Security Policy, Infrastructure and Transport, and Health.

The sentiment classification used binary labels: positive and negative. Positive sentiment refers to supportive, solution-oriented language. Negative sentiment denotes critical or evaluative statements, such as disapproval of policies or actors, and highlights conflicts or failures. Although we use the term negative, it reflects critical tone rather than emotional negativity, aligning with common terminology in machine learning \citep{hofmann2024hate}. As truly neutral speeches are rare, they were excluded. Sentiment and topic analyses used the same sample.

Following these procedures, we curated a labeled dataset of 557 speeches for topic classification and 445 speeches for sentiment classification. The difference in sample size is due to the greater complexity of the multi-class topic task, which required additional labeled examples to ensure robust model training. Table~\ref{tab:sample_data_entry} presents an exemplary data entry.

\begin{table}[h]
    \centering
    \begin{tabularx}{\textwidth}{p{4.6cm}X}
        \toprule
        \textbf{Metadata} & \textbf{Speech} \\
        \midrule
        \begin{tabular}[t]{lp{2.6cm}}
            \textbf{U-id} & 5284 \\
            \textbf{Session-id} & 19/125 \\
            \textbf{Date} & 2019-11-08 \\
            \textbf{Speaker} & Karsten Hilse \\
            \textbf{Party} & AfD \\
            \textbf{Position} & Abgeordneter \\
            \midrule
            \textbf{Topic} & 1 (Environment)\\
            \textbf{Sentiment} & -1 (negative)\\
        \end{tabular}
        &
        Frau Präsidentin! [...] Mit dem Klimaschutzgesetz und dem Emissionshandelsgesetz wird ein weiteres Bürokratiemonster geschaffen, das Bürgerinnen und Bürger finanziell belastet. Die Kosten werden auf die Verbraucher abgewälzt, ohne dass die Wirkung dieser Maßnahmen belegt wäre. [...] Dieses Gesetz basiert auf einer ideologisch motivierten Hypothese und dient letztlich der Transformation unserer Gesellschaft in eine ökosozialistische Diktatur. \\
        \bottomrule
    \end{tabularx}
    \caption{Sample data entry with metadata, assigned topic and sentiment, and speech excerpt (abridged)}
    \label{tab:sample_data_entry}
\end{table}

\subsection{Model Development and Validation}
\label{sec:models}

Separate models were developed for sentiment and topic classification, each trained and evaluated on the manually labeled dataset. Given the distinct nature of these two classification tasks, we applied different strategies tailored to their specific requirements.

\emph{Topic classification} was approached as a multi-class task, where speeches could be assigned to a single category. This required a more semantically rich representation. A custom Word2Vec embedding model was trained on the whole corpus of parliamentary speeches, enabling the capture of domain-specific language patterns \citep{lilleberg_support_2015}. These embeddings outperformed generic pre-trained alternatives. Several classifiers were evaluated, including logistic regression, XGBoost, and a Bagging ensemble with a Support Vector Machine (SVM) base estimator. To obtain an overall score, the area under the receiver operating characteristic curve (AUROC) values were averaged over all topic classes.

\emph{Sentiment classification} was perfomed using Term Frequency-Inverse Document Frequency (TF-IDF) vectorization based on unigrams to transform the speeches into numerical feature representations. This frequency-based method was selected due to its efficiency and strong performance in text classification tasks with limited label complexity \citep{rosenberger2025careerbert}. Several classifiers were evaluated, including Random Forest \citep{breiman2001random}, XGBoost \citep{chen2016xgboost}, and logistic regression.

\emph{Model Evaluation.} All models were trained and tested using an 80:20 train-test split to ensure robustness and generalizability. Performance was evaluated on the test set using three metrics: accuracy, F1-score, and the AUROC. AUROC quantifies a classifier’s ability to distinguish between classes across all threshold levels, with a value of 1.0 indicating perfect separation and 0.5 representing random guessing \citep{fawcett2006introduction}. AUROC served as the primary evaluation metric, as it provides a threshold-independent measure of model performance -- especially relevant when comparing classifiers with different decision thresholds, where accuracy and F1-score can vary substantially \citep{huang2005using}. Accuracy and F1-score complemented this by capturing overall correctness and the balance between precision and recall, respectively.

To assess the \textit{generalizability} of our machine learning models beyond their original training domain, we conducted a small validation study using political speech transcripts from the Austrian National Council (Nationalrat). A manually labeled sample of 80 Austrian speeches served as ground truth to test both topic and sentiment classification.

\subsection{Application to Field Data}
\label{sec:field_data}

The best-performing models, evaluated on the test set, were applied to the full text corpus. Using the predicted labels, we analyzed relative topic distribution over time, party-specific sentiment trends, and sentiment variation by topic within parties. These models provide the analytical foundation for answering the research question through structured, large-scale classification of thematic and sentimental dimensions in parliamentary discourse.

\section{Results}
\label{sec:results}

\subsection{Model Results}
\label{sec:model_results}

Building on the model development outlined in Section~\ref{sec:models}, both the sentiment and topic classification models achieved strong performance. 
Table~\ref{tab:classification_results} presents a overview of the evaluated classifiers for both tasks. The results highlight the relative strengths of each approach in terms of accuracy, F1-score, and AUROC, providing the basis for model selection in each case.

\begin{table}[H]
    \centering
    \setlength{\tabcolsep}{0pt}
    \begin{tabular*}{\textwidth}{@{\extracolsep{\fill}} l S[table-format=1.2] S[table-format=1.2] S[table-format=1.2] l S[table-format=1.2] S[table-format=1.2] S[table-format=1.2] @{}}
        \toprule
        \multicolumn{4}{@{}c}{Topic Classification} & \multicolumn{4}{c@{}}{Sentiment Classification} \\
        \cmidrule{1-4} \cmidrule{5-8}
        Model & {Accuracy} & {F1} & {AUROC} & Model & {Accuracy} & {F1} & {AUROC} \\
        \midrule
        Logistic Regression & 0.62 & 0.62 & 0.90 & Logistic Regression & 0.76 & 0.76 & 0.85 \\
        XGBoost              & 0.63 & 0.63 & 0.90 & XGBoost              & 0.71 & 0.70 & 0.78 \\
        Bagging + SVM        & 0.68 & 0.68 & 0.94 & Random Forest        & 0.85 & 0.85 & 0.89 \\
        \bottomrule
    \end{tabular*}
    \caption{Performance of evaluated classifiers for topic and sentiment classification}
    \label{tab:classification_results}
\end{table}

For \textit{topic classification}, the best-performing model was the Bagging classifier combined with custom Word2Vec embeddings. It achieved an accuracy and F1-score of 0.68. AUROC analysis further supports the model’s effectiveness, with an average AUROC of 0.94 across the six topic classes, indicating strong separability even in the presence of thematic overlap.

For \textit{sentiment classification}, the Random Forest classifier reached an accuracy and F1-score of 0.85. The AUROC value of 0.89 further confirms the model’s ability to robustly distinguish between positive and negative sentiments.

Additionally, both models showed reasonable \textit{generalizability} to Austrian parliamentary data. The topic classifier achieved an AUROC of 0.84, down from 0.94 in the original domain, while the sentiment classifier reached 0.78, compared to 0.89.

Despite maintaining decent discriminative power, both models -- especially the sentiment classifier -- showed lower accuracy and F1-scores. Sentiment accuracy dropped to 0.49, indicating difficulties with threshold calibration in the new context. A closer look revealed a clear asymmetry in misclassifications: genuinely positive speeches were often labeled as negative, while no negative speeches were predicted as positive. This may reflect a distinct tone in Austrian parliamentary discourse, where different expressions are used.

\subsection{Field Data Trends}
\label{sec:Field Data Trends}

This section presents the empirical results derived from applying the classification models to the full Bundestag dataset. The analysis directly addresses the research question introduced earlier, with a particular focus on the three subquestions: (1) identifying dominant topics and their variation across parties and time, (2) examining shifts in sentiment in response to political role changes, and (3) analyzing how specific topics are framed emotionally by different parties. The results are presented in the form of visual analyses based on the enriched parliamentary speech dataset described in Section~\ref{sec:field_data}, providing an in-depth view of discourse patterns in the German Bundestag.

\begin{figure}[h] 
    \centering 
    \includegraphics[width=\textwidth]{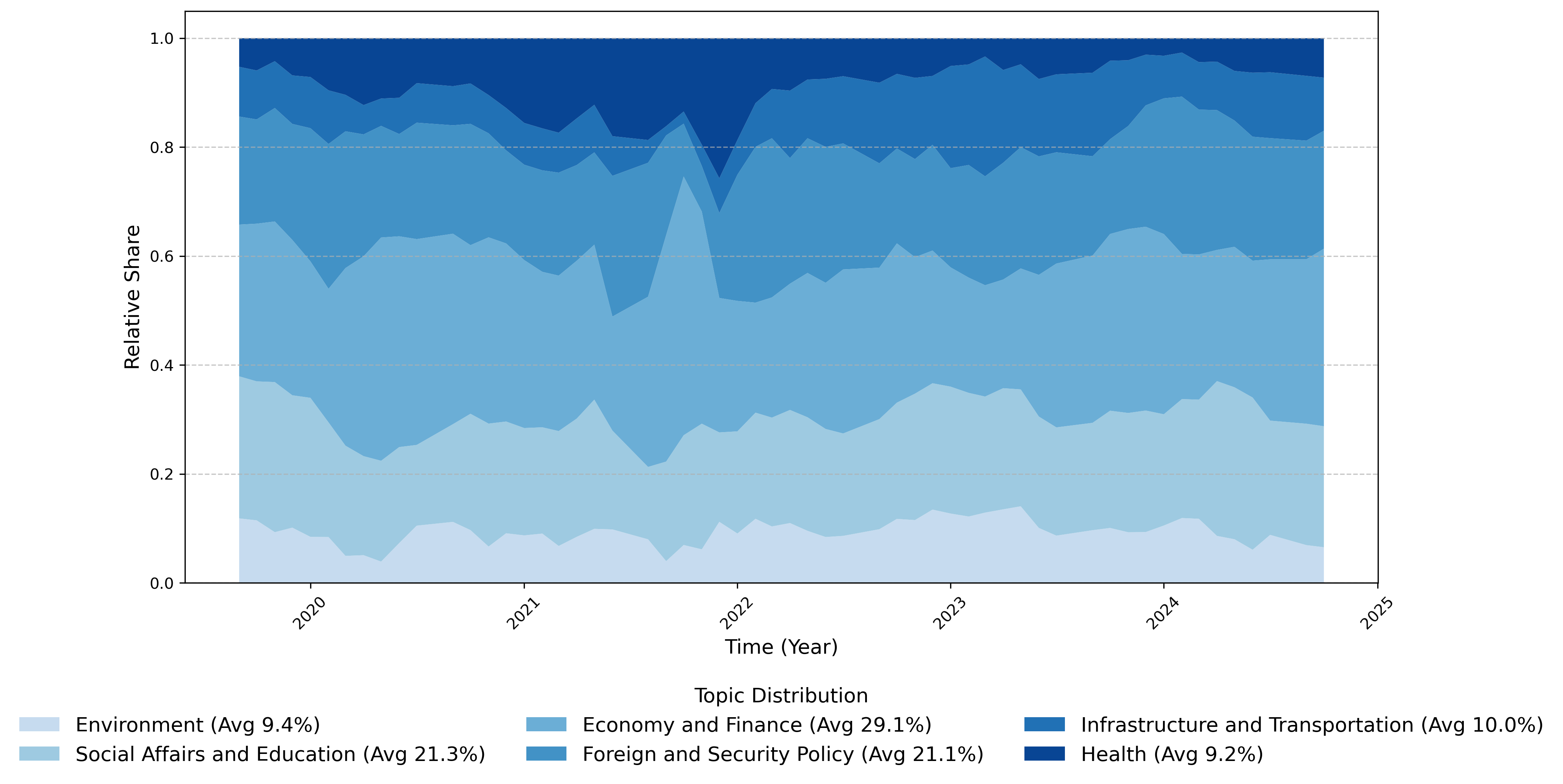} 
    \caption{Topic distribution over time} 
    \label{fig:topic_distribution} 
\end{figure}

\emph{Topic Distribution Over Time.} Figure \ref{fig:topic_distribution} illustrates the distribution of political discussions in the Bundestag across different topics over time. From this analysis, it becomes evident that three main topics, Social Affairs and Education, Economy and Finance, and Foreign and Security Policy, together make up around 70\% of the discussions. These three topics dominate the political debates in the German Parliament, reflecting their central importance in the political landscape. In contrast, topics such as Environment, Health, and Infrastructure and Transportation each account for roughly 10\% of the total discussions, suggesting that they occupy a smaller but still relevant space in the political dialogue. The discussion of Health, in particular, showed a notable fluctuation. During the COVID-19 pandemic, Health-related topics were discussed much more frequently than after the end of the pandemic in early 2022. This shift highlights the substantial impact that the pandemic had on the political agenda. Furthermore, towards the end of 2021, Economy and Finance saw a marked increase in discussions. This surge can be attributed to the formation of the new coalition government (coalition between Sozialdemokratische Partei Deutschlands (SPD), Bündnis 90/Die Grünen and Freie Demokratische Partei (FDP)), which faced substantial challenges related to budget planning. The government's budget had to be passed under the constraints of the debt brake, which had been temporarily suspended during the COVID years \citep{connolly_fear_2023}. This budget was later declared unconstitutional by the Federal Constitutional Court \citep{BVerfG_2BvF1-22_2023-11-15}, which may have further contributed to the heightened discussions around Economic and Financial matters at that time.

\begin{figure}[h] \centering \includegraphics[width=\textwidth]{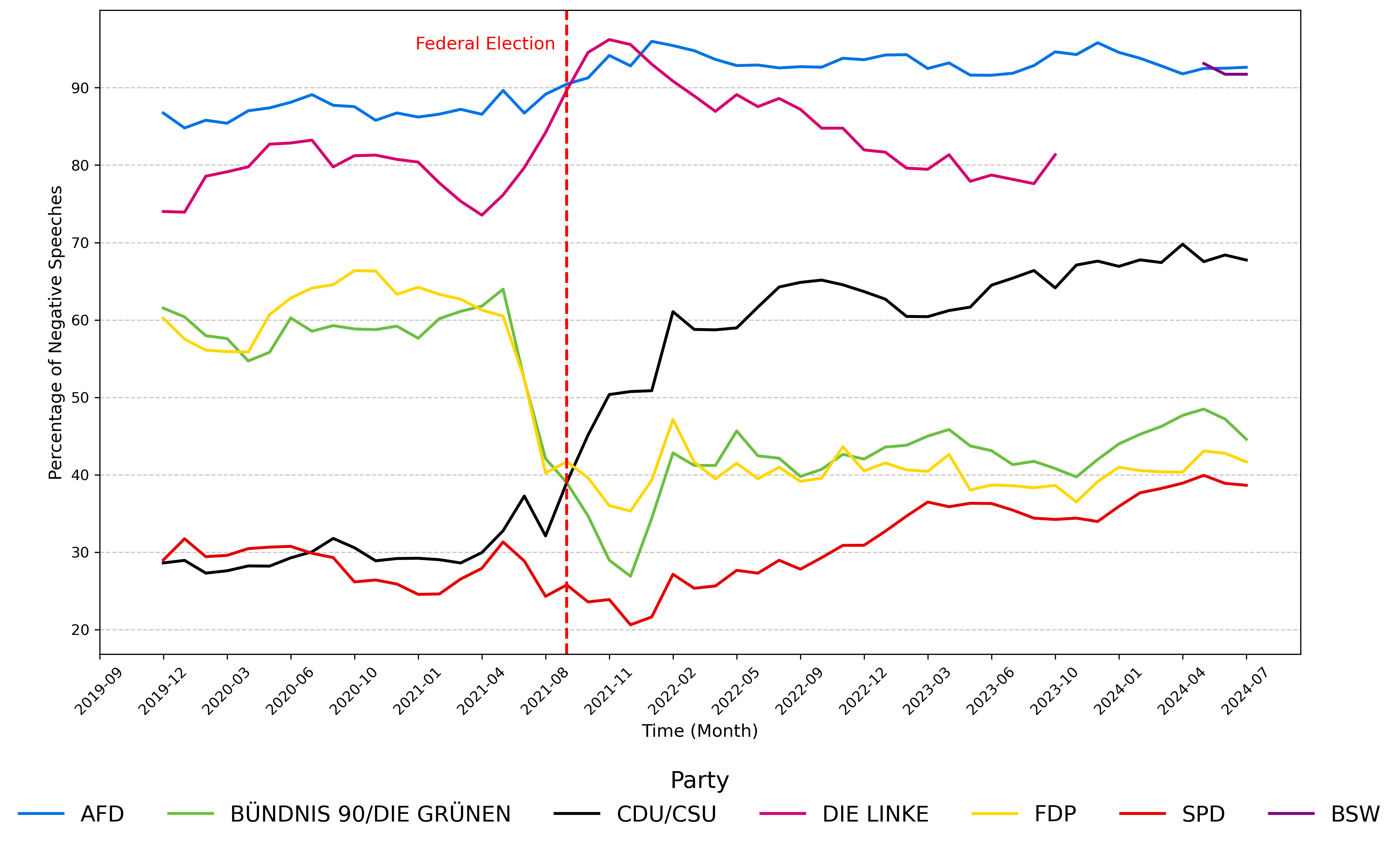} \caption{Percentage of negative speeches over time by party} \label{fig:sentiment_distribution} \end{figure}

\emph{Sentiment Distribution Over Time.} Figure \ref{fig:sentiment_distribution} presents the relative share of negatively labeled speeches per party. This analysis breaks down the distribution of negative sentiments, with each party being represented by its own curve. This visualization provides insight into how the sentiment in speeches has evolved over time for each political party, allowing for a comparative assessment of how parties have shifted in terms of negative speech over the period of observation. One of the most notable observations is the shift in sentiment following the September 2021 elections. Prior to the election, CDU/CSU, as part of the governing Coalition, exhibited a relatively low share of negative speeches. However, after moving into opposition, this share increased sharply. Conversely, when FDP and Bündnis 90/Die Grünen joined the governing coalition, their proportion of critical statements fell markedly. This pattern suggests that parties in government generally express fewer negative sentiments, likely reflecting their perception of their own policies in a more favorable light. In contrast, opposition parties tend to take a more critical stance, resulting in a higher proportion of negative speeches. A comparison between the two governing coalitions further supports this observation. During the 19th legislative period (2017–2021), the governing coalition consisted of CDU/CSU and SPD. During this time, approximately 30\% of speeches from coalition parties carried a negative sentiment. In contrast, the subsequent coalition -- comprising SPD, Bündnis 90/Die Grünen, and FDP -- governing during the 20th legislative period (from 2021), exhibited a notably higher share of negative speeches of around 40\%. This may indicate that this coalition was more critical of its own policies than its predecessor, reflecting a greater degree of internal tension or dissatisfaction within the alliance. These inner conflicts eventually culminated in the premature dissolution of the coalition in November 2024 \citep{meakem_german_2024}. Another key observation is that parties on the political extremes tend to express more negative sentiments. The AfD, widely regarded as the most right-wing party in the German Bundestag, consistently delivers over 90\% of its speeches with a negative tone. Similarly, Die Linke, considered the most left-wing party, uses a negative tone about 80\% of the time. This high level of negativity not only reflects the parties' positions at the extreme end of the political spectrum, but also underlines their role as sharp critics of the ruling government.

\begin{figure}[h] 
    \centering 
    \includegraphics[width=0.9\textwidth]{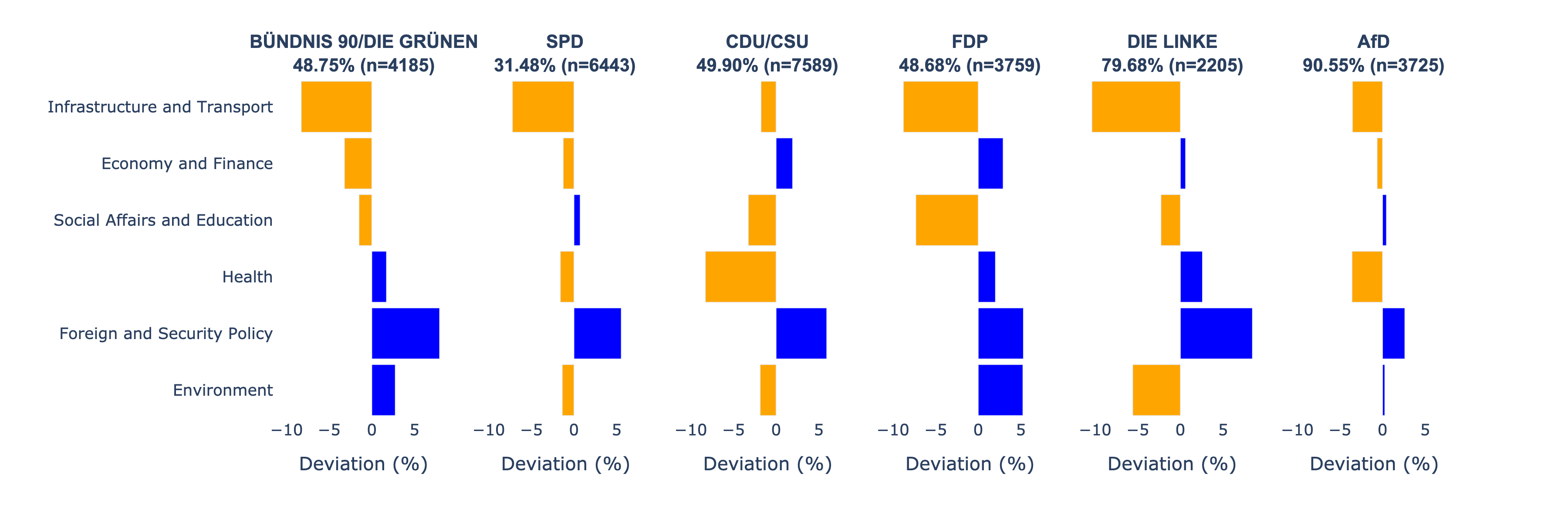} \caption{Deviation sentiment distribution per topic and party} \label{fig:topic_party_time} 
\end{figure}

\emph{Sentiment Distribution per Topic and Party.}
Figure~\ref{fig:topic_party_time} combines both models to conduct a joint analysis. The top part displays the average share of negatively labeled speeches for each party (with total counts in parentheses). Below that, deviations in sentiment per topic are visualized: a positive deviation (blue) indicates that the share of negative speeches for a given topic is higher than the party’s own average across all topics, while a negative deviation (orange) indicates a lower share. Importantly, these deviations are always calculated relative to each party’s individual baseline—not the overall Bundestag average. This enables a relative within-party comparison, showing how critically a party discusses specific policy areas compared to its typical tone.

One key observation is that Infrastructure and Transportation is evaluated more positively across all parties relative to their respective baselines, suggesting lower political contention in this area. In contrast, Foreign and Security Policy tends to be discussed more negatively, especially by opposition parties. This heightened negativity may reflect dissatisfaction with the current political direction in this domain. However, it is important to note that these sentiment deviations do not necessarily reflect topic-specific critiques, but may instead indicate broader discontent with the political status quo—particularly among opposition actors.


Another notable finding is the thematic focus of individual parties. CDU/CSU and FDP, traditionally associated with market-oriented policies, show higher negativity in the area of Economy and Finance. SPD places emphasis on Social and Education topics, while Bündnis 90/Die Grünen dedicate more attention to Environment. Although the frequency of topics is not depicted here, these patterns correspond to well-known party priorities reflected in broader analyses. Interestingly, the intensity of criticism tends to increase within parties’ core issue areas. This suggests that parties use sentiment not just to evaluate, but to strategically frame their priorities and underline the need for political action. Overall, these findings highlight the strategic nature of political discourse in the Bundestag: parties selectively focus on topics and adjust the tone of their rhetoric to align with their broader political objectives.

\section{Discussion}
\label{sec:discussion}

Our study presents an approach for automated classification of parliamentary speeches, contributing both methodologically and empirically to the computational analysis of political discourse. First, a dataset of speeches from the German Bundestag was collected, manually annotated, and preprocessed with respect to domain-specific challenges.

Second, the study demonstrates that machine learning methods, if adapted to the linguistic and structural characteristics of political speech, can yield meaningful insights. The successful application of different classification strategies tailored to sentiment and topic detection shows the practical value of combining established NLP techniques with contextual knowledge. The approach of developing domain-specific classifiers offers a potentially transferable framework for analyzing other structured text domains, such as political communication, media analysis, or institutional reporting.

Third, the application of the trained models to real-world Bundestag data enables large-scale empirical analysis and reveals systematic patterns in parliamentary discourse. Debates are dominated by a few key policy areas such as Economy and Finance, Social and Education, and Foreign and Security Policy. Sentiment patterns differ substantially between government and opposition parties, with opposition parties consistently expressing more negative sentiment overall. Additionally, parties tend to adopt a particularly critical stance within their own core issue areas, suggesting a strategic use of negativity to emphasize political priorities. The resulting dataset, enriched based on model predictions, offers new possibilities for tracking discourse dynamics over time and across party, thereby expanding the analytical scope for political science and discourse studies.

From a theoretical perspective, our study shows that machine learning can be effectively applied to structured political text when domain-specific nuances are considered. Unlike general NLP tasks, classifying parliamentary speech demands sensitivity to institutional language, formal rhetoric, and the hybrid nature of political expression, which shifts between critique and consensus. Custom embeddings and task-specific feature engineering were essential for capturing these aspects.

In contrast to previous work that has relied on unsupervised methods for the analysis of parliamentary texts, such as the study by \citet{rauh2018validating}, our study adopts a supervised approach grounded in domain-specific manual annotations. This enables a precise classification of sentiment and topics, particularly in complex or institutionally embedded speech segments. By combining the advantages of manual labeling with scalable machine learning techniques, the approach presented extends and refines existing methodologies.

Despite the promising results, several limitations must be acknowledged. First, model performance, though high in relative terms, still leaves room for improvement. Particularly in topic classification, overlapping themes and ambiguous phrasing contributed to misclassifications, suggesting potential gains from multi-label classification or more advanced machine learning models. Second, the manual labeling process, despite internal validation and cross-checking, remains subjective, especially where speeches touch upon multiple thematic areas or where sentiment is expressed in subtle rhetorical forms. 

Third, the size of the labeled training data remains limited. While the overall corpus comprises over 28,000 speeches, only a small subset of entries was used for supervised model training. This could constrain model robustness, particularly for underrepresented topics. Future work should focus on expanding the labeled dataset -- either through crowdsourcing, active learning, or semi-supervised approaches -- to enhance both coverage and performance. To facilitate further research, all datasets and code used in this study are publicly available.\footnote{\href{https://github.com/ptzlukas/german-parliamentary-speech-classifier}{https://github.com/ptzlukas/german-parliamentary-speech-classifier}}

In conclusion, our study demonstrates the feasibility and value of applying machine learning to large-scale political speech data. Our work contributes to a deeper understanding of parliamentary discourse and offers a framework for future machine learning–based political science research. Beyond methodological contributions, our findings answer our research questions by revealing empirical patterns in the evolution of political topics, sentiment dynamics over time, and party roles. They also reveal the emotional framing of different areas. These results highlight the value of computational techniques in generating evidence-based insights into parliamentary communication.


\bibliographystyle{agsm}
\bibliography{literature}

\end{document}